\documentclass[letterpaper, 10 pt, conference]{ieeeconf}  

\IEEEoverridecommandlockouts                              

\newtheorem{proposition}{Proposition}
\newtheorem{corollary}{Corollary}

\newcommand{\ph}[1]{\textcolor{black}{#1}}

\usepackage{cite}
\usepackage{amsmath,amssymb,amsfonts}
\usepackage{graphicx}
\usepackage{textcomp}
\usepackage[table]{xcolor}
\usepackage{booktabs}
\usepackage{multirow}
\usepackage{makecell}
\usepackage{array}
\usepackage{hyperref}
\usepackage{adjustbox}

\title{What Limits mmWave Human-State Sensing for Robot Perception?}

\author{
Shuntian Zheng, Guangming Wang, Hongyang He, Shuai He, Yu GUAN*\\
shuntian.zheng,
Hongyang.He@warwick.ac.uk,
gw462@cam.ac.uk,
hs19951021@bupt.edu.cn,
yu.guan@warwick.ac.uk
}

\begin{document}

\maketitle
\thispagestyle{empty}
\pagestyle{empty}


\begin{abstract}

Assistive robots and search-and-rescue platforms require reliable human-state sensing in smoke, darkness, and privacy-sensitive environments, where cameras and LiDAR may be unreliable. 
Millimeter-wave (mmWave) radar offers an alternative that works without ambient light, penetrates smoke, and does not capture visual appearance. 
However, the compact arrays that fit robot size, power, and cost budgets have limited apertures, producing joint-localization errors on the order of decimeters at a few meters, large enough to compromise the safety margins and tracking accuracy a robot depends on. 
Existing pipelines reduce pose error through frame fusion, multiple viewpoints, return selection, or stronger estimators, but examine one technique at a time, without relating its effect to a shared measurement limit or separating gains from added information from gains from estimator completion. 
We address this gap with a framework that models the human and environment as one time-varying scattering field and states a Cramér–Rao-type error floor for joint localization as a function of bandwidth, aperture, signal-to-noise ratio, fused frame count, phase coherence, and visibility of retained returns. 
We then use existing frame-fusion, multi-viewpoint, return-selection, and pose-estimation methods as matched probes of this floor rather than as competing pipelines. 
Under the evaluated conditions, experiments on three datasets and a live deployment show these strategies move joint-localization error in the predicted directions, for example lowering mean per-joint position error from about 97 to 65 mm with a second viewpoint, while a stronger estimator mainly lowers overall error by completing joints the measurement cannot resolve. 
These results give the robotics community a principled basis for deciding whether to prioritize observation design or estimator capacity when deploying reliable mmWave human sensing. 

\end{abstract}


\section{Introduction}
\label{sec:intro}

\begin{figure}[t]
  \centering
  \includegraphics[width=\linewidth]{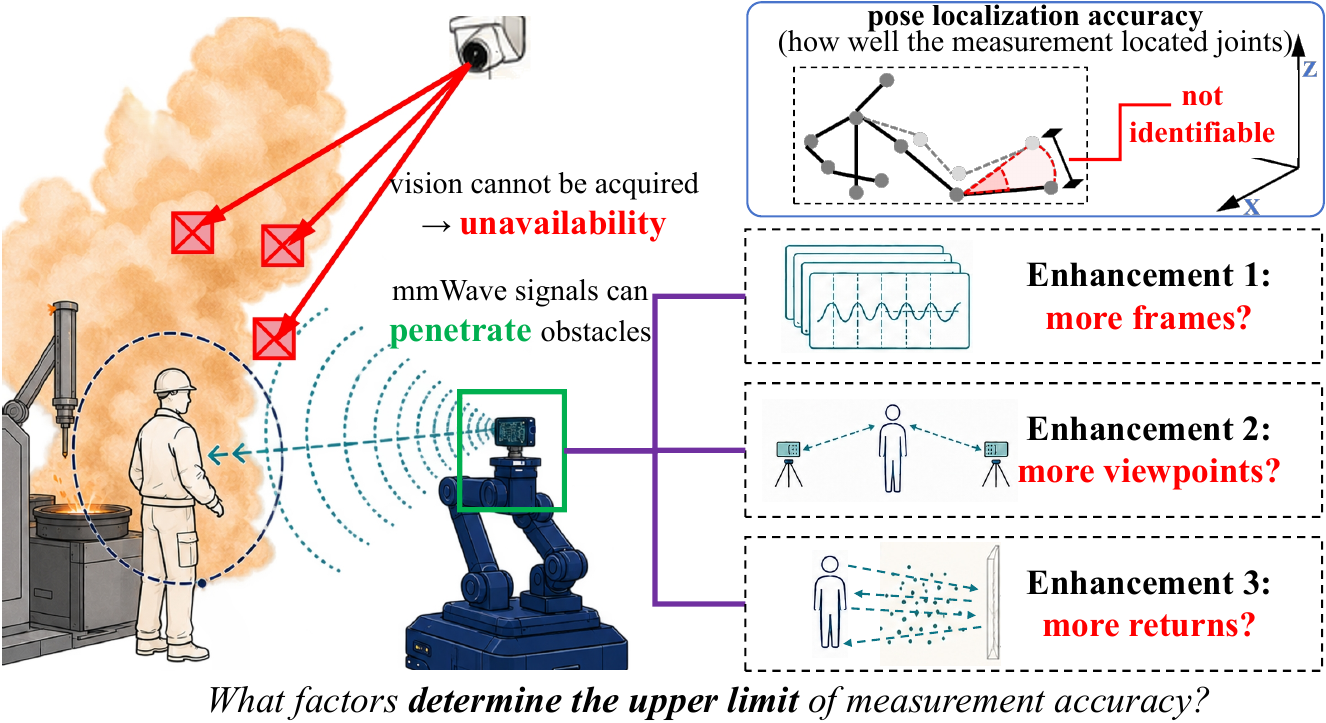}
    \caption{Smoke in an industrial workspace blocks a camera, whereas a robot-mounted mmWave can still observe the worker; some joints nevertheless remain not identifiable. The right-hand callouts label the common front-end choices examined in this paper, namely frame fusion, viewpoint, and return selection, and we ask which sensing quantities set the localization limit.}

  \label{fig:teaser}
\end{figure}

Human-aware robots need reliable estimates of where a person is and how the body is articulated, so that standoff distances, tracking covariance, and contact or avoidance plans remain trustworthy~\cite{fan2024enhancing,yang2025mmdear}.
Cameras and LiDAR supply dense geometry under favorable optics, but steam, smoke, darkness, and privacy-sensitive indoor layouts routinely break that assumption, as in the factory scene of Fig.~\ref{fig:teaser}.

Millimeter-wave (mmWave) radar is a complementary front end for these settings because radio returns remain usable without ambient light, penetrate fog and steam, and do not archive visual appearance~\cite{zheng2025person}.
This promise is strongest on the compact single-chip arrays that fit on-robot size, power, and cost budgets.
Yet those arrays resolve human joints far more coarsely than optical range sensors: a representative $3\mathrm{T}{\times}4\mathrm{R}$ front end already yields near-$14^{\circ}$ azimuth beamwidth~\cite{rahman2024mmvr}, which maps to decimeter-scale lateral uncertainty at a few meters and can erode interactive margins even when a downstream network outputs a plausible skeleton.
Throughout, \textbf{pose localization accuracy} therefore denotes the accuracy with which the radar measurement itself can locate body joints, as distinct from the visual fidelity of an estimated mesh or pose sequence; joints the measurement does not constrain remain \emph{not identifiable} (Fig.~\ref{fig:teaser}, top right).

This paper asks: \textbf{How accurately can mmWave locate human joints, and which sensing quantities set that limit?}

A growing literature already reduces pose error without answering that question.
Temporal and spatial aggregation fuse successive frames or additional viewpoints to enlarge the observation and average noise~\cite{zheng2026learn,choi2025mvdoppler}.
Learned human-state estimators map sparse radar features onto dense joint or mesh outputs~\cite{lai2026radarllm,luan2024diffusion}.
A further line retains static environmental reflections instead of discarding them as clutter~\cite{zhou2026rise}; we treat this choice as return selection, parallel to multi-frame and multi-view enrichment.
These families typically keep a conventional preprocessing chain fixed, report gains under a single reconstruction metric, and leave unspecified both how accurately the measurement can locate the body and which sensing quantities remain responsible for the residual gap.
Empirical improvements therefore accumulate as pipeline scores, without a shared criterion for whether the next gain should come from more frames, additional viewpoints, a different return policy, or a larger estimator (Fig.~\ref{fig:teaser}, right).

We attribute this bottleneck to a \textbf{missing measurement characterization}: enhancements alter different stages of the sensing chain without a shared \textbf{error floor} on pose localization accuracy.
Three structural mismatches follow.
First, multi-frame and multi-view aggregation often proceeds under implicit assumptions that phase stays aligned across the fused frames (coherence) or that the combined viewpoints are geometrically calibrated, so apparent gains may reflect favorable phase alignment or geometry rather than a controlled information increase.
Second, reconstruction metrics do not separate \emph{measurement information}, structure already constrained by the observation, from \emph{estimator completion} of coordinates the radar never rendered identifiable.
Third, return selection changes which scatterers enter the observation, yet rarely quantifies the resulting information about the human state once the unknown environmental geometry is marginalized, that is, averaged out as a nuisance rather than estimated.
Absent a common formalism that separates these effects, existing results do not state an error floor on human-joint localization imposed by the radar measurement itself, independent of which enhancement is applied.

We therefore close that missing measurement characterization instead of adding another reconstruction network.
We model human and environment as one time-varying scattering field and state a Cram\'er--Rao-type error floor on pose localization, so that the sensing quantities altered by existing enhancements can be compared under one reference.
Those enhancements then serve as controlled probes of the floor, which we label \textsc{Fuse}, \textsc{View}, \textsc{Est}, and \textsc{Vis}, in one-to-one correspondence with the mismatches above:
\textsc{Fuse} and \textsc{View} isolate temporal and spatial aggregation;
\textsc{Est} separates measurement information from estimator completion;
and \textsc{Vis} tests when retaining weak or reflected returns is informative about the human state.

Concretely, under matched conditions and varying one probe at a time on complementary public benchmarks:
\textsc{Fuse} lowers error on the joints the measurement itself renders identifiable with a log--log slope of \ph{$-0.28$} in the fused frame count, while mean per-joint position error (MPJPE) over all joints improves more slowly, consistent with estimator completion;
\textsc{View} lowers MPJPE from \ph{$96.6$} to \ph{$64.8\,\mathrm{mm}$} once a second viewpoint is added;
and \textsc{Vis} lowers error on identifiable joints by \ph{$10.4\%$} and MPJPE by \ph{$9.3\%$}, with a larger margin in high-multipath scenes and no resolved effect under near-stationary motion or severe phase drift.
A live deployment reproduces the multi-frame and visibility trends under real sensor noise, including indoor captures with up to three concurrent subjects.
These probe-level trends anchor the following contributions.

The main contributions are as follows:

(C1)~\textbf{Measurement-limit diagnosis.}
We identify the core limitation of lightweight mmWave human-state sensing for robots as a missing measurement characterization of pose localization accuracy, and we separate this sensing limit from estimator completion.

(C2)~\textbf{Measurement-imposed error floor.}
We state a Cram\'er--Rao-type error floor on human-joint localization as an explicit function of the sensing quantities that front-end choices alter, providing a common reference for comparing those choices; the experiments vary the subset those choices can change at deployment time, namely fused frame count, coherence, viewpoint count, and visibility.

(C3)~\textbf{Factor probes via existing enhancements.}
We recover multi-frame, multi-view, return-selection, and estimator pipelines as matched probes, yielding falsifiable, factor-level predictions without a new end-to-end architecture.

(C4)~\textbf{Public and live validation.}
We test these predictions on complementary datasets and on a live prototype with up to three concurrent subjects, quantifying where measurement-type error follows the predicted trends, most directly through the fusion slope of \ph{$-0.28$} on identifiable joints, and where estimator completion accounts for the remaining gain.

\section{Related Work}
\label{sec:related_work}

%
This section organizes prior work along the three lines implicated by the mismatches above, and asks, for each line, whether it separates measurement from estimator completion or supplies a factor-level sensing limit for robot front ends.

\subsection{Human-State Estimation under Fixed Observations}

Early mmWave systems showed that radar returns can support localization, activity recognition, and pose estimation without visual appearance, and benchmarks such as mvdoppler~\cite{choi2025mvdoppler}, MM-Fi~\cite{yang2023mm}, and mmRadPose~\cite{mueller2025radproposer} then enabled supervised evaluation of 3D pose and body reconstruction across subjects, actions, and environments.
Subsequent estimators, including point-based encoders~\cite{yang2025spatial}, temporal transformers~\cite{zheng2026learn}, and diffusion densifiers~\cite{fan2024diffusion}, improve articulated recovery for human--robot interaction and privacy-preserving reconstruction.
These systems primarily instantiate the downstream estimation stage of the pipeline formalized in Sec.~\ref{sec:formulation}: they predict the human state from an observation fixed by conventional preprocessing, often retaining only returns attributed directly to the body, and do not ask how accurately that measurement can locate joints~\cite{zhou2026rise}, nor how the limit changes when the observation itself is redesigned.
They advance estimator completion while leaving the measurement-imposed error floor unspecified, which is exactly the distinction Corollary~\ref{cor:c2} later formalizes.

\subsection{Multi-Frame and Multi-View Aggregation}

A second line improves the observation by changing temporal or spatial aggregation before estimation: temporal fusion enlarges the observation window and averages noise, whereas multi-view processing adds angular baselines or complementary viewpoints.
Recent systems make the operator roles especially clear: PULSE supports single- and multi-frame RAD inference~\cite{zheng2026doppler}, milliMamba aggregates multi-frame heatmaps under specular sparsity~\cite{kini2025millimamba}, and MVDoppler-Pose fuses multi-view range--Doppler streams~\cite{choi2025mvdoppler}.
These designs instantiate the temporal or multi-view fusion stage formalized in Sec.~\ref{sec:formulation} and empirically probe fused frame count, phase coherence, and viewpoint geometry, yet reported gains are judged by reconstruction error under a fixed front-end convention, without stating the corresponding scaling of the joint-localization error floor or verifying the coherence and calibration assumptions that make aggregation informative.

\subsection{Return Selection, Visibility, and Clutter Conventions}

A third line changes which scatterers enter the observation: conventional clutter suppression discards static or weakly dynamic returns, whereas recent work retains static environmental reflections or human-induced multipath to enrich fine-grained sensing, including RISE's treatment of multipath as geometric signal rather than noise~\cite{zhou2026rise}.
We treat this choice as return selection that instantiates the front-end selection stage formalized in Sec.~\ref{sec:formulation} and probes visibility, parallel in intent to multi-frame and multi-view enrichment rather than as a separate sensing objective.
Current analyses often target layout recovery or propagation modeling and rarely quantify the marginal information about the human state after the unknown environmental geometry is accounted for, so retaining reflections cannot be compared with suppressing them inside a shared error floor.


\begin{figure*}[t]
\centering
\includegraphics[width=\linewidth]{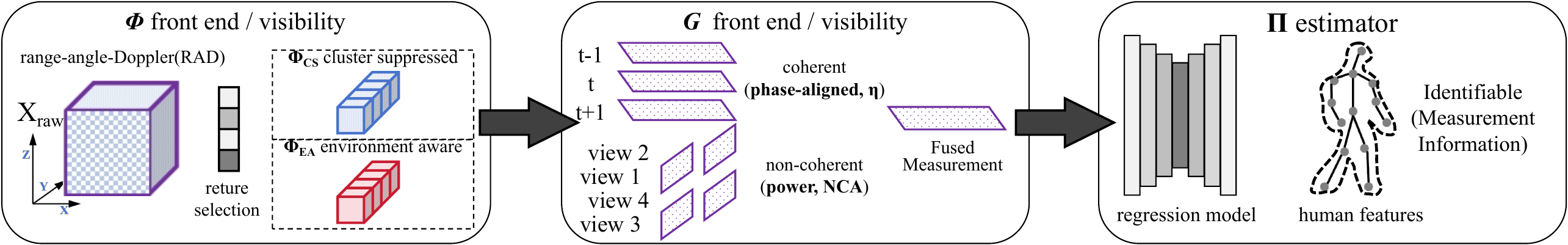}
\caption{Operator view of pipelines. $\Phi$ applies return selection to a raw range--angle--Doppler cube $\mathbf{x}_{\mathrm{raw}}$, yielding a clutter-suppressed ($\Phi_{\mathrm{CS}}$) or environment-aware ($\Phi_{\mathrm{EA}}$) observation. $\mathcal{G}$ aggregates frames, coherently or non-coherently, and optional viewpoints into a fused measurement. $\Pi$ maps that measurement to a pose; labeled joints are those identifiable from measurement information. Each operator is later used as a factor probe under one error floor.}
\label{fig:framework}
\end{figure*}

\section{Signal Model and Problem Formulation}
\label{sec:formulation}

Section~\ref{sec:related_work} shows that existing enhancements modify different stages of a single sensing chain, yet are compared only by pipeline scores.
This section defines the objects those stages act on: the physical superposition of human and environmental returns, the estimand--nuisance split, the retained visibility, and the three operators in Fig.~\ref{fig:framework}, so that Sec.~\ref{sec:theory} can place a single error floor on the chain.

\textbf{Forward model.}
In an FMCW front end, every scatterer in the scene, on the body or in the environment, contributes a path to the same baseband sample.
Let $\boldsymbol{\theta}_H(t)\in\mathbb{R}^{d_H}$ be the human state, comprising joint locations and short-horizon kinematics when needed, and let $\boldsymbol{\theta}_E$ denote unknown environmental geometry.
The sample at virtual antenna $n$, fast time $\tau$, and slow time $t_s$ is
\begin{equation}
\small
\label{eq:forward}
s_n(\tau,t_s)
=
\sum_{p\in\mathcal{P}_{\mathrm{all}}(t_s)}
\alpha_p(t_s)\,e^{j\phi_p(\tau,t_s)}
+
w_n(\tau,t_s),
\end{equation}
where $\mathcal{P}_{\mathrm{all}}$ indexes all active paths, whether direct or reflected, $\alpha_p$ is a complex amplitude, and $w_n$ is noise.
After dechirp, the phase of path $p$ is
\begin{equation}
\small
\label{eq:phase}
\phi_p(\tau,t_s)
=
\frac{4\pi R_p}{\lambda}
+
\frac{4\pi B R_p\,\tau}{c\,T_c}
+
\frac{4\pi v_p t_s}{\lambda}
+
\frac{2\pi n d\sin\theta_p}{\lambda},
\end{equation}
where $R_p$ is the equivalent range of path $p$, that is, half its total propagation length, so that reflected paths enter with their unfolded geometry; $B$ is chirp bandwidth, $T_c$ chirp duration, $v_p$ radial velocity, $\theta_p$ azimuth, $\lambda$ the carrier wavelength, and $d$ the virtual-element spacing.
Range is encoded in the fast-time beat, Doppler in slow time across chirps, and angle in the array manifold.
Keeping or discarding a path is therefore an observation-design choice rather than a property of the physics.
Single-frame Fourier resolutions $\Delta R=c/(2B)$ and $\Delta\theta\approx\lambda/(D\cos\theta)$, with $D$ the array aperture, already expose bandwidth and aperture as sensing quantities, but do not give a joint-localization error floor once frames are fused and returns are re-selected.

\textbf{Estimand, nuisance, and visibility.}
The robot task estimates $\boldsymbol{\theta}_H$, whereas $\boldsymbol{\theta}_E$ is a nuisance that nonetheless shapes every reflected return.
Let the \emph{visibility} $\mathcal{V}$ be the set of paths a front end retains: a clutter-suppressed (CS) convention keeps the directly visible set $\mathcal{V}_{\mathrm{CS}}$, whereas an environment-aware (EA) convention additionally retains reflected returns, so that $\mathcal{V}_{\mathrm{EA}}\supseteq\mathcal{V}_{\mathrm{CS}}$~\cite{yang2025spatial}.
Stacking sufficient statistics up to bounce order $M$ yields
\begin{equation}
\small
\label{eq:stacked}
\mathbf{z}_t
=
\begin{bmatrix}
\mathbf{z}_t^{(0)\top}
&
\cdots
&
\mathbf{z}_t^{(M)\top}
\end{bmatrix}^{\!\top}
=
\mathbf{h}\!\left(\boldsymbol{\theta}_H(t),\boldsymbol{\theta}_E\right)
+
\mathbf{n}_t,
\end{equation}
where $\mathbf{z}_t^{(m)}$ aggregates $m$-bounce features, $\mathbf{h}$ is induced by Eqs.~\eqref{eq:forward}--\eqref{eq:phase}, and $\mathbf{n}_t$ is noise.
The two conventions differ only in the visibility that truncates $\mathbf{z}_t$, so comparing them on equal terms requires marginalizing the unknown $\boldsymbol{\theta}_E$, which Sec.~\ref{sec:theory} carries out explicitly.

\textbf{Operator decomposition.}
Any practical pipeline admits
\begin{equation}
\small
\label{eq:operators}
\hat{\boldsymbol{\theta}}_H
=
\Pi\circ\mathcal{G}\circ\Phi\!\left(\mathbf{x}_{\mathrm{raw}}\right),
\end{equation}
where $\Phi$ selects RAD features and retained returns from $\mathbf{x}_{\mathrm{raw}}$, $\mathcal{G}$ performs temporal or multi-view fusion into a fused measurement, and $\Pi$ is a learned or geometric estimator (Fig.~\ref{fig:framework}).
The CS and EA conventions instantiate $\Phi$ as $\Phi_{\mathrm{CS}}$ and $\Phi_{\mathrm{EA}}$, with hard static--dynamic splits a further variant, and thereby probe visibility $\mathcal{V}$ under \textsc{Vis}; coherent, non-coherent, and two-view fusion instantiate $\mathcal{G}$ and thereby probe fused frame count $K$, coherence $\eta$, and viewpoint geometry under \textsc{Fuse} and \textsc{View}; pose heads instantiate $\Pi$ under \textsc{Est}.
An explicit error floor on this chain remains.

\section{Measurement-Imposed Error Floor and Its Governing Factors}
\label{sec:theory}

We now place a lower bound on the error with which the observation in Fig.~\ref{fig:framework} can locate joints, and map each governing factor to a probe of $\Phi$, $\mathcal{G}$, or $\Pi$.
The identities below are standard Cram\'er--Rao scalings under stated local assumptions, not a claim of a new tight 3D bound; their role is to fix the factor ordering that experiments must test.

\textbf{Single-frame angular limit.}
To expose how array geometry and SNR set the floor before fusion, the \emph{effective aperture}:
\begin{equation}
\small
\label{eq:deff}
D_{\mathrm{eff}}
=
D_{\mathrm{phys}}
+
\sum_{c\in\mathcal{C}_{\mathrm{arr}}}\Delta D_c,
\end{equation}
where $D_{\mathrm{phys}}$ is the physical array length $D$ of Sec.~\ref{sec:formulation} and $\Delta D_c$ is the extra baseline contributed by a coplanar, array-calibrated element or colocated view $c\in\mathcal{C}_{\mathrm{arr}}$.
Two cases are excluded: reflected environmental paths, which enlarge $\mathcal{V}$ and enter $\boldsymbol{\theta}_E$ rather than the aperture, and widely separated radars, which add geometric diversity rather than length of a single linear array, so that \textsc{View} tests viewpoint count and not an additive $D_{\mathrm{eff}}$ term.
With path amplitude $\alpha$, noise variance $\sigma_n^{2}$, and $\mathrm{SNR}=|\alpha|^2/\sigma_n^{2}$, the standard array Cram\'er--Rao result~\cite{van2004detection} gives a \emph{variance} bound that scales:
\begin{equation}
\small
\label{eq:crb1}
\mathrm{CRB}_1(\theta)
\;\propto\;
\frac{1}{D_{\mathrm{eff}}^{2}\cos^{2}\theta\cdot\mathrm{SNR}}.
\end{equation}
Larger $D_{\mathrm{eff}}$ or SNR tightens $\mathrm{CRB}_1(\theta)$; RMSE then scales as the square root of this variance.

\textbf{Multi-frame scaling.}
Fusion changes that floor through frame count and phase quality.
Intuitively, frames whose phase stays aligned can be summed coherently before squaring, so their signal amplitudes add while independent noise partially cancels; frames combined only in power average out noise more slowly but do not require that alignment.
Let $K$ be the number of fused frames and $\eta=\exp(-\sigma_\phi^{2}/2)\in(0,1]$ the coherence factor for residual phase error variance $\sigma_\phi^{2}$ after compensation.
Since $\sigma_\phi^{2}$ accumulates over the fusion window, $\eta$ is a property of that window rather than a constant, and is written $\eta(K)$.
\begin{proposition}[Multi-frame angular limits]
\label{prop:t2}
Assume a local Gaussian observation and residual phase error that is well modeled by $\eta$.
Coherent stacking multiplies SNR by $K\eta$, so the variance CRB scales as $1/(K\eta)$.
Independent power-domain accumulation multiplies SNR by $\sqrt{K}$ in the low-to-moderate SNR regime of non-coherent integration~\cite{richards2005fundamentals}, which is the regime used here (median \ph{$14\,\mathrm{dB}$}):
\begin{equation}
\small
\label{eq:t2coh}
\mathrm{CRB}_{K}^{(\mathrm{coh})}(\theta)
\approx
\frac{\mathrm{CRB}_1(\theta)}{K\eta},
\qquad
\mathrm{CRB}_{K}^{(\mathrm{NCA})}(\theta)
\approx
\frac{\mathrm{CRB}_1(\theta)}{\sqrt{K}}.
\end{equation}
The corresponding RMSEs therefore scale as $(K\eta)^{-1/2}$ and $K^{-1/4}$.
Comparing RMSEs, coherent fusion is preferable when $K>1/\eta^{2}$.
\end{proposition}

As $\eta$ drops, the crossover $K>1/\eta^{2}$ moves to larger $K$, and this ordering of $K$ and $\eta$ is what \textsc{Fuse} tests.
Because the realized scaling is governed by $K\eta(K)$ rather than by $K$ alone, a fused frame count buys less than $(K)^{-1/2}$ whenever longer windows degrade phase alignment, so \textsc{Fuse} reports $\hat{\eta}$ at every $K$ instead of assuming it fixed.

\textbf{From CRB to joint localization.}
The angle and range bounds above become a statement about pose localization accuracy once they are mapped to body joints.
Since the array reports angle and range rather than Cartesian position directly, a fixed angular error corresponds to a larger positional error the farther the joint is from the sensor, so converting the bounds above into millimeters requires accounting for that range.
Let $\mathcal{U}(\boldsymbol{\theta}_H,\mathcal{V})$ be the joints with nonsingular local Fisher information under visibility $\mathcal{V}$, and write joint $j\in\mathcal{U}$ in polar form $(R_j,\theta_j)$ in the array's azimuth--range plane.
\begin{proposition}[Joint-localization error floor]
\label{prop:t3}
Under a local Gaussian, single-scattering-center model in the azimuth--range plane,
\begin{equation}
\small
\label{eq:t3}
\mathbb{E}\,\|\hat{\mathbf{q}}_j-\mathbf{q}_j\|^{2}
\;\gtrsim\;
R_j^{2}\,\mathrm{CRB}_{K}^{(\theta)}(\theta_j)
+
\mathrm{CRB}_{K}^{(R)}(R_j),
\end{equation}
where $\mathrm{CRB}_{K}^{(\theta)}$ follows Proposition~\ref{prop:t2} and $\mathrm{CRB}_{K}^{(R)}$ is the range CRB with the same SNR multiplier, so both terms inherit the coherent $K\eta$ or non-coherent $\sqrt{K}$ scaling, with $B$ setting the single-frame range scale.
\end{proposition}
The $R_j^{2}$ term converts azimuth uncertainty into Cartesian meters at range.
The sum is the azimuth--range diagonal form used as the Cartesian reference: compact $3\mathrm{T}{\times}4\mathrm{R}$ arrays have a much coarser elevation beam, which the closed form omits, and range--angle cross-covariance terms of the full Fisher matrix are dropped.
The floor is therefore this 2D reference rather than a tight 3D bound, which is how the Cartesian CRB of Sec.~\ref{sec:experiments} is reported.
For $j\notin\mathcal{U}$, no finite measurement-imposed floor exists: a network may still output that joint, but such completion adds no measurement information.

\textbf{Visibility after marginalizing $\boldsymbol{\theta}_E$.}
Return selection enlarges $\mathcal{V}$, but the retained reflected paths depend on the unknown layout $\boldsymbol{\theta}_E$, so their contribution must be assessed after $\boldsymbol{\theta}_E$ is marginalized.
Under additive Gaussian noise $\mathbf{n}\sim\mathcal{N}(0,\mathbf{R}_n)$ on Eq.~\eqref{eq:stacked}, write $\boldsymbol{\psi}=[\boldsymbol{\theta}_H^{\top},\boldsymbol{\theta}_E^{\top}]^{\top}$ and let $\mathbf{H}=\partial\mathbf{h}/\partial\boldsymbol{\theta}_H$, $\mathbf{E}=\partial\mathbf{h}/\partial\boldsymbol{\theta}_E$.
An EA observation then has joint Fisher information
\begin{equation}
\small
\label{eq:fim}
\mathbf{J}_{\mathrm{EA}}(\boldsymbol{\psi})
=
\begin{bmatrix}
\mathbf{J}_{HH} & \mathbf{J}_{HE} \\
\mathbf{J}_{EH} & \mathbf{J}_{EE}
\end{bmatrix},
\qquad
\mathbf{J}_{AB}=\mathbf{A}^{\top}\mathbf{R}_n^{-1}\mathbf{B},
\end{equation}
where the block indices $A,B\in\{H,E\}$ select the corresponding Jacobian $\mathbf{H}$ or $\mathbf{E}$, so that $\mathbf{J}_{EH}=\mathbf{J}_{HE}^{\top}$.
If $\mathbf{R}_n\succ0$, $\mathbf{h}$ is locally differentiable, and $\mathbf{J}_{EE}$ is invertible, marginalizing $\boldsymbol{\theta}_E$ via the standard Schur-complement identity for nuisance parameters~\cite{van2004detection} yields
\begin{equation}
\small
\label{eq:schur}
\tilde{\mathbf{J}}_{\mathrm{EA}}(\boldsymbol{\theta}_H)
=
\mathbf{J}_{HH}-\mathbf{J}_{HE}\mathbf{J}_{EE}^{-1}\mathbf{J}_{EH}.
\end{equation}
A CS front end instead applies a projection $P_{\mathrm{CS}}$ to $\mathbf{z}_t$, which leaves $\mathbf{J}_{\mathrm{CS}}=\mathbf{H}^{\top}P_{\mathrm{CS}}^{\top}(P_{\mathrm{CS}}\mathbf{R}_nP_{\mathrm{CS}}^{\top})^{-1}P_{\mathrm{CS}}\mathbf{H}$.
\begin{proposition}[Information ordering of retained returns]
\label{prop:t1}
Assume the Gaussian, differentiability, and invertibility conditions above, and in addition that $\mathbf{R}_n$ is block diagonal across bounce orders, so that the direct block $\mathbf{z}_t^{(0)}$ is uncorrelated with the retained reflected blocks.
Then the marginalized information about $\boldsymbol{\theta}_H$ after retaining reflected (or low-Doppler) returns satisfies
\begin{equation}
\small
\label{eq:t1}
\tilde{\mathbf{J}}_{\mathrm{EA}}(\boldsymbol{\theta}_H)
\;\succeq\;
\mathbf{J}_{\mathrm{CS}}(\boldsymbol{\theta}_H).
\end{equation}
Equality holds when the retained block is collinear with the direct block or is unidentifiable after eliminating $\boldsymbol{\theta}_E$.
\end{proposition}
\emph{Proof sketch.}
Partition the Jacobians by bounce order as $\mathbf{H}=[\mathbf{H}_0^{\top},\mathbf{H}_r^{\top}]^{\top}$ and $\mathbf{E}=[\mathbf{0}^{\top},\mathbf{E}_r^{\top}]^{\top}$, the latter because the direct block does not depend on $\boldsymbol{\theta}_E$.
Block diagonality of $\mathbf{R}_n=\mathrm{blkdiag}(\mathbf{R}_0,\mathbf{R}_r)$ then splits Eq.~\eqref{eq:fim} additively and leaves $\mathbf{J}_{HE}$ and $\mathbf{J}_{EE}$ supported on the retained block alone, while $P_{\mathrm{CS}}$ retains exactly $\mathbf{z}_t^{(0)}$, giving $\mathbf{J}_{\mathrm{CS}}=\mathbf{H}_0^{\top}\mathbf{R}_0^{-1}\mathbf{H}_0$.
Substituting into Eq.~\eqref{eq:schur} yields
$\tilde{\mathbf{J}}_{\mathrm{EA}}-\mathbf{J}_{\mathrm{CS}}
=\mathbf{H}_r^{\top}\mathbf{R}_r^{-1/2}(\mathbf{I}-\mathbf{P}_E)\mathbf{R}_r^{-1/2}\mathbf{H}_r$,
where $\mathbf{P}_E$ is the orthogonal projection onto the column space of $\mathbf{R}_r^{-1/2}\mathbf{E}_r$.
Since $\mathbf{I}-\mathbf{P}_E\succeq0$, Eq.~\eqref{eq:t1} follows, with equality exactly when $\mathbf{R}_r^{-1/2}\mathbf{H}_r$ lies in that column space, which is the collinear or unidentifiable case.

Eq.~\eqref{eq:t1} is the information-theoretic content of the \textsc{Vis} probe, including static--dynamic splits that drop human-coupled low-Doppler returns.
The inequality does not assert that retaining reflections must reduce MPJPE under every motion regime; it only ranks Fisher information when the regularity conditions hold.
The same information-theoretic view also bounds what a stronger downstream estimator can add once the observation itself is fixed:

\begin{corollary}[Measurement versus estimator completion]
\label{cor:c2}
For fixed $\Phi$ and $\mathcal{G}$, no downstream $\Pi$ can increase Fisher information in the observation; it may still lower reconstruction error by completing joints outside $\mathcal{U}$.
\end{corollary}
The first clause is the standard property that post-processing cannot increase Fisher information~\cite{van2004detection}; what is specific here is its use as an evaluation criterion, since only \emph{measurement information} is governed by the error floor, whereas \emph{estimator completion} explains score gains that do not raise sensing-limited accuracy (\textsc{Est}).

Table~\ref{tab:factor-protocol} in Sec.~\ref{sec:experiments} maps these statements onto matched operator contrasts.
The local predictions to be tested are:
(\textbf{H1}) moving from $\Phi_{\mathrm{CS}}$ to $\Phi_{\mathrm{EA}}$ raises the marginalized information about $\boldsymbol{\theta}_H$, so error on $\mathcal{U}$ falls, and it falls further where multipath is richer;
(\textbf{H5}) within $\Phi_{\mathrm{EA}}$, a unified observation improves on a hard static--dynamic split only while the retained low-Doppler block stays identifiable, so the advantage vanishes once it does not;
(\textbf{H2}, \textbf{H3}) under coherent fusion, RMSE on $\mathcal{U}$ scales closer to $(K\eta)^{-1/2}$ than does MPJPE, and the advantage of coherent over non-coherent fusion shrinks as $\eta$ drops;
(\textbf{H4}) adding a complementary calibrated viewpoint reduces Cartesian localization error on $\mathcal{U}$ as well as overall, under matched $\Phi$ and $\Pi$.
Sec.~\ref{sec:discussion} records the settings in which these regularity conditions fail, without overturning the factor ordering above.

\section{Experiments: Factor-Level Validation}
\label{sec:experiments}

%
This section tests predictions with matched operator contrasts: each subsection isolates one sensing quantity while holding the remaining operators in Eq.~\eqref{eq:operators} fixed.
The question is whether the observed trend agrees in direction with the corresponding prediction, not which pipeline scores highest.

\subsection{Experimental Setup}
\label{sec:exp-setup}

\begin{table*}[t]
\caption{Factor-to-probe protocol. Each row isolates one sensing quantity while holding the remaining operators in Eq.~\eqref{eq:operators} fixed.}
\label{tab:factor-protocol}
\centering
\begin{adjustbox}{width=\linewidth,center}
\setlength{\tabcolsep}{4pt}
\begin{tabular}{p{2.6cm}p{7.2cm}p{3.8cm}p{1.3cm}}
\Xhline{1.2pt}
\toprule
\textbf{Factor} & \textbf{Probe instantiation} & \textbf{Controlled operator change} & \textbf{Prediction} \\
\midrule
Frames and coherence
& PULSE single-/multi-frame; milliMamba temporal aggregation
& $\mathcal{G}_{\mathrm{coh}}$, $\mathcal{G}_{\mathrm{NCA}}$; vary $K$
& \textbf{H2}, \textbf{H3} \\
\rowcolor[HTML]{EFEFEF}
Viewpoint geometry
& MVDoppler-Pose two-view fusion~\cite{choi2025mvdoppler}
& Single view versus two complementary views; matched $\Phi,\Pi$
& \textbf{H4} \\
Visibility
& RISE-style multipath / static-reflection retention
& $\Phi_{\mathrm{CS}}$ versus $\Phi_{\mathrm{EA}}$; split versus unified observation
& \textbf{H1}, \textbf{H5} \\
\rowcolor[HTML]{EFEFEF}
Estimator capacity
& Frozen heatmap / PC backbone versus stronger $\Pi_{\mathrm{dense}}$
& Freeze $\Phi,\mathcal{G}$; vary only $\Pi$
& Corollary~\ref{cor:c2} \\
\bottomrule
\Xhline{1.2pt}
\end{tabular}
\end{adjustbox}
\end{table*}

Introducing a dedicated network for each factor would confound architecture capacity with the sensing effect under study.
We therefore reuse existing enhancements as probes of $\Phi$ and $\mathcal{G}$ (Table~\ref{tab:factor-protocol}) and freeze $\Pi$ across the Fuse, View, and Vis contrasts, following Corollary~\ref{cor:c2}.
Within the heatmap family we freeze a shared RAD pose backbone $\Pi_{\mathrm{hm}}$ on mmRadPose; within the point-cloud family we freeze an MM-Fi-compatible backbone $\Pi_{\mathrm{pc}}$.
%
Only the Est contrast replaces the frozen head with a stronger completion module $\Pi_{\mathrm{dense}}$.
Heads are not re-tuned to a particular $\Phi$ or $\mathcal{G}$ setting.
As instantiated here, $\Phi_{\mathrm{CS}}$ applies hard CFAR suppression, $\Phi_{\mathrm{EA}}$ retains low-threshold reflected features, and $\mathcal{G}_{\mathrm{NCA}}$ / $\mathcal{G}_{\mathrm{coh}}$ realize non-coherent and coherent aggregation.


The probes are taken from representative published baselines: milliMamba~\cite{kini2025millimamba}, PULSE~\cite{zheng2026doppler}, and RISE~\cite{zhou2026rise}; they are modified only as needed to isolate one operator on a shared backbone (Table~\ref{tab:factor-protocol}).
RISE is the method we found that explicitly retains previously discarded static reflections; we keep that front end and append an MLP that maps its human-state features onto each corpus's skeleton, so Vis reports MPJPE on the same joints as the other probes.
Where those sources report 3D MPJPE on the same corpus, we accept a probe as a faithful re-instantiation only if it stays within \ph{$5\%$} of the published single-modality score; 

\paragraph{Datasets}
No single corpus isolates every governing factor under one representation.
\textbf{mmRadPose} provides OptiTrack-aligned 3D skeletons and radar cubes from which RAD heatmaps can be formed~\cite{mueller2025radproposer}; it is the primary heatmap corpus for Fuse, Vis, and Est.
\textbf{MM-Fi} is the point-cloud corpus used to check that Fuse and Vis signs survive a change of representation and a public cross-environment split~\cite{yang2023mm}.
\textbf{MVDoppler-Pose} supplies synchronized two-radar 3D walking poses~\cite{choi2025mvdoppler} and is used only for View.

\paragraph{Metrics}
We report mean per-joint position error (MPJPE, mm) and, to separate measurement information from estimator completion, RMSE on a \emph{fixed} identifiable set $\mathcal{U}$: joints whose radar return exceeds a detection threshold.
One rule builds $\mathcal{U}$ once per corpus at the richest observation of that probe ($K{=}\ph{10}$ with $\Phi_{\mathrm{EA}}$ on mmRadPose, two views on MVDoppler-Pose, the matched operating point on MM-Fi) and freezes it across conditions; the rule uses return statistics rather than estimator error, so RMSE on $\mathcal{U}$ indexes measurement information and overall MPJPE additionally reflects completion.
View also reports azimuth RMSE.
A Cartesian CRB reference on $\mathcal{U}$ follows Proposition~\ref{prop:t3} at the measured $\hat{\eta}(K)$, with calibrated $B{=}\ph{4}\,\mathrm{GHz}$, $D_{\mathrm{eff}}{\approx}\ph{7}\lambda$, median SNR $\ph{14}\,\mathrm{dB}$, and median range \ph{$2.8\,\mathrm{m}$}, and is compared only with RMSE on $\mathcal{U}$.
Errors are means over \ph{5} seeds; a difference is called \emph{resolved} only if a sequence-level paired bootstrap (\ph{2000} resamples) has a $95\%$ interval excluding zero, about \ph{$1.8\,\mathrm{mm}$} in MPJPE and \ph{$1.5\,\mathrm{mm}$} on $\mathcal{U}$.

\subsection{Multi-Frame Fusion and Coherence under \textsc{Fuse}}
\label{sec:exp-h2}

Proposition~\ref{prop:t2} predicts two signatures for fusion: RMSE on $\mathcal{U}$ should track $(K\eta)^{-1/2}$ more closely than all-joint MPJPE, and the advantage of coherent over non-coherent aggregation should shrink as $\eta$ drops.
On mmRadPose we vary only $K$ under $\mathcal{G}_{\mathrm{coh}}$ versus $\mathcal{G}_{\mathrm{NCA}}$, with $\Phi$ and $\Pi$ frozen (Table~\ref{tab:fuse-k} and Fig.~\ref{fig:factor-validation}(a)).
The window coherence $\hat{\eta}$ is taken from the residual phase variance of stable scatterers after compensation, and does not use pose labels.

Coherent RMSE on $\mathcal{U}$ has log--log slope \ph{$-0.28$}, against \ph{$-0.17$} for MPJPE, which remains larger at every $K$.
A fixed-$\eta$ reading of Proposition~\ref{prop:t2} would give $-0.5$; substituting the measured $\hat{\eta}(K)$ into Eq.~\eqref{eq:t3} instead gives a floor slope of \ph{$-0.30$}, and RMSE on $\mathcal{U}$ stays above that curve throughout the sweep.
Non-coherent fusion, which does not depend on $\eta$, has slope \ph{$-0.19$} against the predicted $-0.25$, so the coherent comparison should be read at that tolerance.
At $K{=}\ph{5}$ the table reports the pooled $\hat{\eta}{=}\ph{0.54}$; splitting the same windows by residual phase into high- and low-coherence halves gives $\ph{0.71}$ and $\ph{0.38}$, and the coherent advantage on $\mathcal{U}$ holds only on the former ($K>1/\eta^{2}$), matching Proposition~\ref{prop:t2}.
The two signatures support (\textbf{H2}, \textbf{H3}): error on identifiable joints follows $K$ and $\eta$ as measurement information, rather than a generic reduction of MPJPE with longer temporal context.

\begin{table}[t]
\caption{Fuse probe on mmRadPose under heatmap $\Phi$ and frozen $\Pi$. Bold marks Coh values that improve over NCA at the same $K$ by a resolved margin (Sec.~\ref{sec:exp-setup}); the bottom row reports the log--log slope of each metric against $K$.}
\label{tab:fuse-k}
\centering
\begin{adjustbox}{width=\linewidth,center}
\setlength{\tabcolsep}{1pt}
\begin{tabular}{l|lccccc}
\Xhline{1.2pt}
\toprule
$\mathcal{G}$ & Metric & $K{=}1$ & $K{=}5$ & $K{=}10$ & $K{=}20$ & $K{=}40$ \\
\midrule
NCA
& RMSE on $\mathcal{U}$
& \ph{126.1}
& \ph{96.0}
& \ph{83.2}
& \ph{72.1}
& \ph{62.5} \\
\rowcolor[HTML]{EFEFEF}
Coh
& RMSE on $\mathcal{U}$
& \ph{126.1}
& \ph{\textbf{80.0}}
& \ph{\textbf{66.2}}
& \ph{\textbf{54.4}}
& \ph{\textbf{44.8}} \\
NCA
& MPJPE
& \ph{132.0}
& \ph{108.5}
& \ph{98.6}
& \ph{92.4}
& \ph{88.0} \\
\rowcolor[HTML]{EFEFEF}
Coh
& MPJPE
& \ph{132.0}
& \ph{\textbf{91.2}}
& \ph{\textbf{78.4}}
& \ph{\textbf{73.6}}
& \ph{\textbf{71.2}} \\
\midrule
Coh
& $\hat{\eta}$
& \ph{--}
& \ph{0.54}
& \ph{0.41}
& \ph{0.31}
& \ph{0.23} \\
\rowcolor[HTML]{EFEFEF}
Coh
& CRB ref.\ on $\mathcal{U}$
& \ph{103.8}
& \ph{64.0}
& \ph{52.0}
& \ph{42.2}
& \ph{34.3} \\
\midrule
Log--log
& \multicolumn{6}{l}{\ph{RMSE$_\mathcal{U}$:NCA$-0.19$, Coh$-0.28$; MPJPE Coh$-0.17$; CRB ref.\ $-0.30$}} \\
\bottomrule
\Xhline{1.2pt}
\end{tabular}
\end{adjustbox}
\end{table}

\subsection{Two-View Geometry under \textsc{View}}
\label{sec:exp-h4}

Proposition~\ref{prop:t3} implies that a complementary viewpoint should reduce Cartesian localization error under matched $\Phi$ and $\Pi$.
This tests geometric diversity rather than an additive $D_{\mathrm{eff}}$ scaling of a single linear array.
On MVDoppler-Pose we compare one radar against two complementary radars using MVDoppler-Pose-style two-view fusion, with $\Phi$ and $\Pi$ held fixed (Fig.~\ref{fig:factor-validation}(b)).
Azimuth RMSE decreases from \ph{$8.4^{\circ}$} to \ph{$5.1^{\circ}$}, RMSE on $\mathcal{U}$ from \ph{$86.9$} to \ph{$56.4\,\mathrm{mm}$} (\ph{$35.1\%$}), and MPJPE from \ph{$96.6$} to \ph{$64.8\,\mathrm{mm}$} (\ph{$32.9\%$}); all three changes are resolved, so error falls on $\mathcal{U}$ as well as overall, which supports \textbf{H4} once the other operators are frozen.

\subsection{Visibility and Return Selection under \textsc{Vis}}
\label{sec:exp-h1}

\begin{figure*}[t]
\centering
\includegraphics[width=\linewidth]{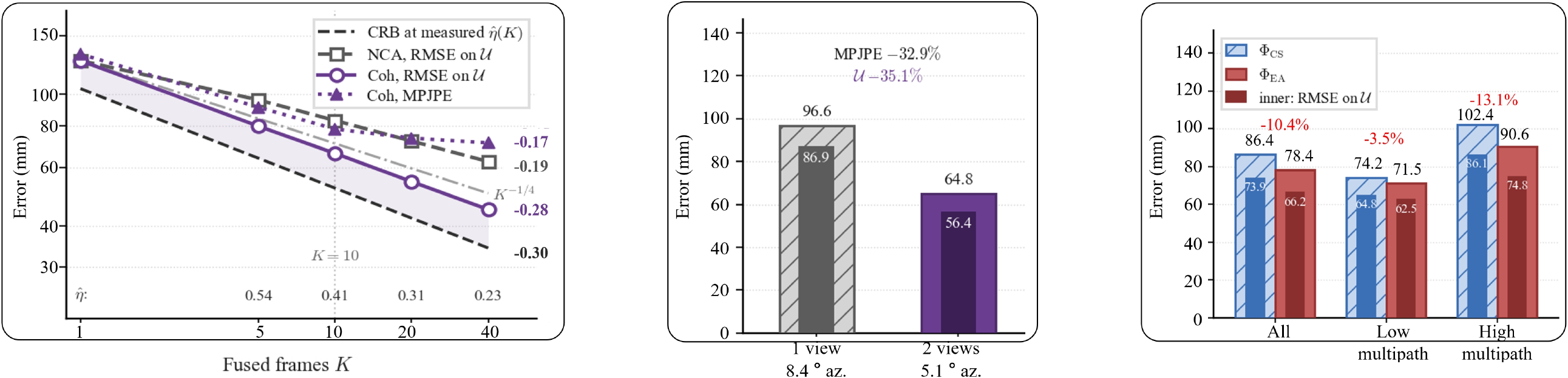}
\caption{Factor-level validation. Each panel changes one operator associated with $K$/$\eta$, viewpoint count, or $\mathcal{V}$, holding the others fixed.
(a)~Fuse: RMSE on $\mathcal{U}$ versus fused frame count $K$, with overall MPJPE, the $K^{-1/4}$ guide, and the CRB reference evaluated at the measured $\hat{\eta}(K)$ printed below the axis; shading marks the remaining distance to that floor.
(b)~View: one versus two complementary radars; outer bars give MPJPE, inner bars RMSE on $\mathcal{U}$, and azimuth RMSE is printed under each tick.
(c)~Vis: clutter-suppressed versus environment-aware return selection under the same inner-bar convention, including the extreme multipath tertiles (the mid tertile is in Table~\ref{tab:vis-main}).}
\label{fig:factor-validation}
\end{figure*}

Proposition~\ref{prop:t1} ranks marginalized Fisher information when an identifiable reflected or low-Doppler block is retained; it does not claim that MPJPE must fall under every motion regime.
On mmRadPose we compare $\Phi_{\mathrm{CS}}$ with a RISE-style reflection-retaining $\Phi_{\mathrm{EA}}$ under matched $\mathcal{G}_{\mathrm{coh}}$ ($K{=}\ph{10}$), the same skeleton MLP, and the same $\Pi$, so only the retention rule changes (Table~\ref{tab:vis-main} and Fig.~\ref{fig:factor-validation}(c)).

Retaining reflected returns lowers RMSE on $\mathcal{U}$ from \ph{$73.9$} to \ph{$66.2\,\mathrm{mm}$} (\ph{$10.4\%$}) and MPJPE from \ph{$86.4$} to \ph{$78.4\,\mathrm{mm}$} (\ph{$9.3\%$}); both changes are resolved.
The drop on $\mathcal{U}$ is larger in the high-multipath tertile (\ph{$13.1\%$}) than in the low-multipath tertile (\ph{$3.5\%$}), which supports \textbf{H1}; the accompanying low-multipath MPJPE change (\ph{$3.6\%$}) is of the same order and is not used as a directional contrast.
Within $\Phi_{\mathrm{EA}}$, replacing a hard static--dynamic split by a unified observation improves RMSE on $\mathcal{U}$ by \ph{$7.2\%$} under moderate motion, whereas the changes under near-stationary micro-Doppler dominance (\ph{$0.8\,\mathrm{mm}$}) and under severe phase drift (\ph{$0.7\,\mathrm{mm}$}, in favor of the split) are not resolved.
This is the pattern \textbf{H5} predicts; the phase-drift subset is also where the block-diagonal noise condition of Proposition~\ref{prop:t1} is least plausible, so we do not claim the information ordering there.

\begin{table}[t]
\caption{Visibility probe under $\mathcal{G}$ and $\Pi$ ($K{=}\ph{10}$). Multipath tertiles and motion subsets contain \ph{N} and \ph{M} sequences, respectively. Bold marks the value in each contrastive pair ($\Phi_{\mathrm{EA}}$ vs.\ $\Phi_{\mathrm{CS}}$; UN vs.\ SD) that differs by a resolved margin.}
\label{tab:vis-main}
\centering
\begin{adjustbox}{width=\linewidth,center}
\setlength{\tabcolsep}{12pt}
\begin{tabular}{lcccc}
\Xhline{1.2pt}
\toprule
& \multicolumn{2}{c}{MPJPE (mm)$\downarrow$} & \multicolumn{2}{c}{RMSE on $\mathcal{U}$ (mm)$\downarrow$} \\
\cmidrule(lr){2-3}\cmidrule(lr){4-5}
Split & $\Phi_{\mathrm{CS}}$ & $\Phi_{\mathrm{EA}}$ & $\Phi_{\mathrm{CS}}$ & $\Phi_{\mathrm{EA}}$ \\
\midrule
All
& \ph{86.4}
& \ph{\textbf{78.4}}
& \ph{73.9}
& \ph{\textbf{66.2}} \\
\rowcolor[HTML]{EFEFEF}
Low multipath
& \ph{74.2}
& \ph{\textbf{71.5}}
& \ph{64.8}
& \ph{\textbf{62.5}} \\
Mid multipath
& \ph{82.6}
& \ph{\textbf{73.1}}
& \ph{70.8}
& \ph{\textbf{61.3}} \\
High multipath
& \ph{102.4}
& \ph{\textbf{90.6}}
& \ph{86.1}
& \ph{\textbf{74.8}} \\
\midrule
\emph{Within $\Phi_{\mathrm{EA}}$} & SD & UN & SD & UN \\
\midrule
Moderate motion
& \ph{84.6}
& \ph{\textbf{78.4}}
& \ph{71.3}
& \ph{\textbf{66.2}} \\
\rowcolor[HTML]{EFEFEF}
Near-stationary
& \ph{82.8}
& \ph{81.2}
& \ph{68.9}
& \ph{68.1} \\
Phase drift
& \ph{88.6}
& \ph{89.4}
& \ph{75.2}
& \ph{75.9} \\
\bottomrule
\Xhline{1.2pt}
\end{tabular}
\end{adjustbox}
\end{table}

\begin{figure}[t]
\centering
\includegraphics[width=0.9271\linewidth]{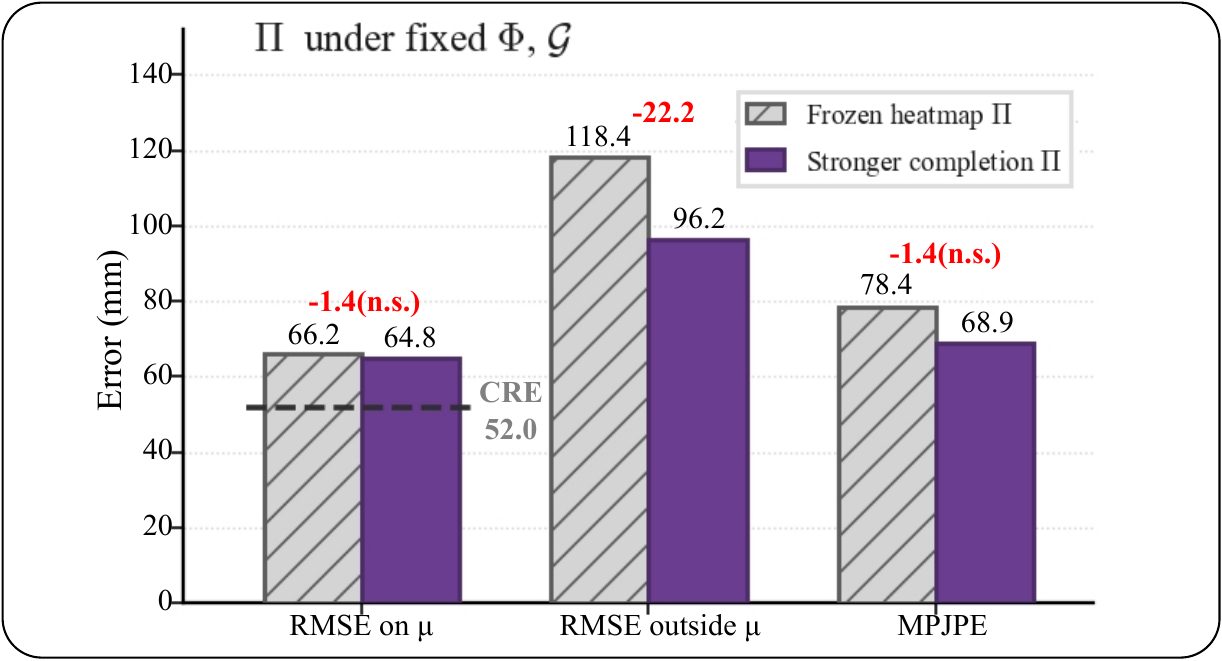}
\caption{Estimator probe under frozen $\Phi$ and $\mathcal{G}$ on mmRadPose ($K{=}\ph{10}$, $\Phi_{\mathrm{EA}}$).
A stronger $\Pi$ reduces overall MPJPE mainly outside $\mathcal{U}$; the change on $\mathcal{U}$ is not resolved, and the CRB reference is unchanged by construction because Eq.~\eqref{eq:t3} does not depend on $\Pi$.}
\label{fig:est-gap}
\end{figure}

\subsection{Estimator Completion versus Measurement Information}
\label{sec:exp-est}

Corollary~\ref{cor:c2} states that, for fixed $\Phi$ and $\mathcal{G}$, a stronger $\Pi$ can lower reconstruction error by completing joints outside $\mathcal{U}$, but it cannot increase measurement Fisher information.
On mmRadPose we freeze $K{=}\ph{10}$, $\Phi_{\mathrm{EA}}$, and the Metrics set $\mathcal{U}$, then swap only $\Pi$ between the frozen heatmap backbone and a stronger PULSE-style completion head (Fig.~\ref{fig:est-gap}).

Overall MPJPE decreases from \ph{$78.4\,\mathrm{mm}$} to \ph{$68.9\,\mathrm{mm}$} (\ph{$12.1\%$} relative), almost entirely outside $\mathcal{U}$: RMSE falls from \ph{$118.4$} to \ph{$96.2\,\mathrm{mm}$} off $\mathcal{U}$, versus \ph{$66.2$} to \ph{$64.8\,\mathrm{mm}$} on $\mathcal{U}$.
The \ph{$1.4\,\mathrm{mm}$} change on $\mathcal{U}$ is not resolved, whereas the \ph{$22.2\,\mathrm{mm}$} change off $\mathcal{U}$ is; this asymmetry, not the aggregate improvement, is what Corollary~\ref{cor:c2} predicts.
The CRB reference on $\mathcal{U}$ is identical under both estimators by construction, since Eq.~\eqref{eq:t3} takes no argument from $\Pi$; the evidential content is that the measured RMSE on $\mathcal{U}$ likewise does not move, leaving the ratio to the floor at about \ph{$1.27$}.
Repeating the swap at $K{=}\ph{40}$ under the same $\Phi_{\mathrm{EA}}$ gives the same split, with MPJPE falling from \ph{$71.2$} to \ph{$63.5\,\mathrm{mm}$} and an unresolved \ph{$1.1\,\mathrm{mm}$} change on $\mathcal{U}$, so the effect is not specific to one operating point.
A point-cloud counterpart on MM-Fi reproduces the same split under the public cross-environment protocol: the Fuse slope is \ph{$-0.22$}, the Vis gain is \ph{$-6.4\%$}, and the stronger $\Pi$ lowers MPJPE from \ph{$112.5$} to \ph{$99.1\,\mathrm{mm}$}, of which \ph{$11.8$} of the \ph{$13.4\,\mathrm{mm}$} arises outside $\mathcal{U}$~\cite{yang2023mm}.
Under the evaluated settings, Fuse, View, and Vis move error in the predicted directions, whereas a stronger estimator narrows residual MPJPE mainly through completion outside $\mathcal{U}$.

\begin{figure*}[h]
\centering
\includegraphics[width=\linewidth]{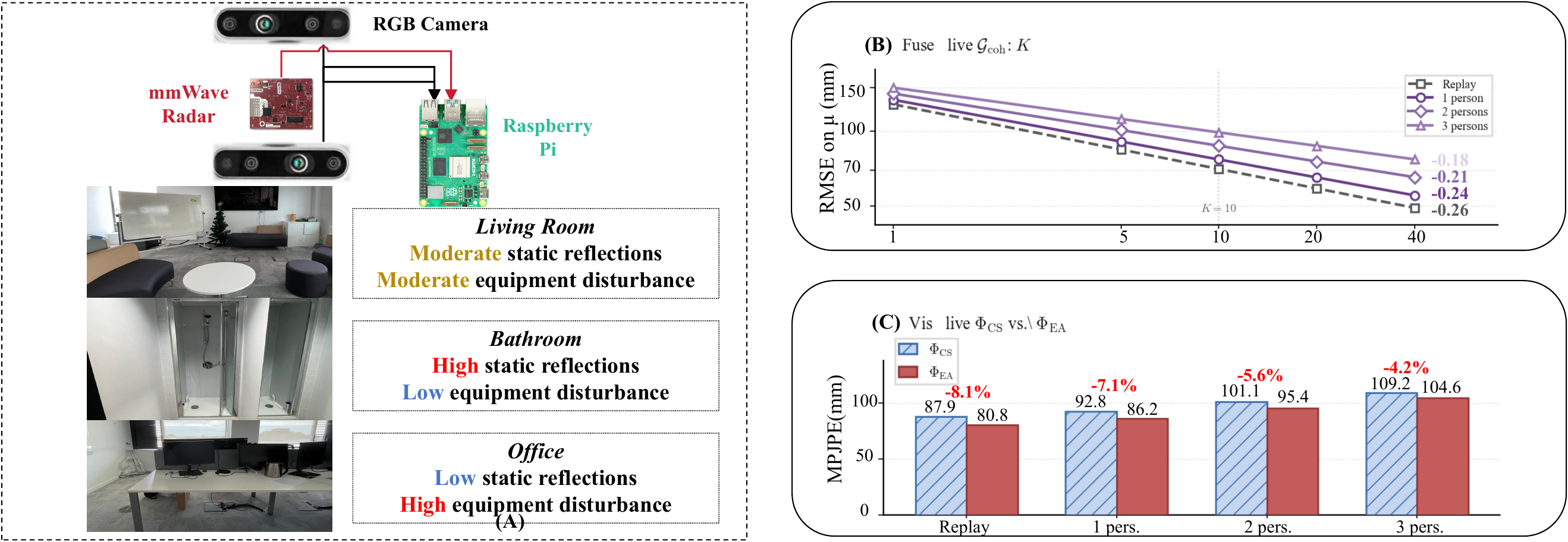}
\caption{Live AWR1843 capture.
(a)~Hardware and three indoor environments; co-located RGB is used only for annotation.
(b)~Coherent RMSE on $\mathcal{U}$ versus fused frame count $K$; log--log slopes remain negative from one to three subjects.
(c)~Clutter-suppressed versus environment-aware MPJPE at the matched operating point.
Dataset-replay is a same-host single-person reference rather than a copy of the public tables.}
\label{fig:proto_scenarios}
\end{figure*}

\section{Real-World Factor Validation}
\label{sec:prototype}

The public contrasts in Sec.~\ref{sec:experiments} are predominantly single-person recordings.
To test whether the Fuse and Vis signs survive real sensor noise and mild multi-person interference, we deploy an AWR1843BOOST radar at $60$--$64\,\mathrm{GHz}$ with co-located RGB used only as an annotation reference, and collect short indoor sequences in an office, a living room, and a bathroom with one to three concurrent subjects under matched $\Phi$ and $\mathcal{G}$ (Fig.~\ref{fig:proto_scenarios}(a)).
Fig.~\ref{fig:proto_scenarios}(b,c) compares these live Fuse and Vis contrasts with a same-host replay on held-out single-person captures from the same rooms, stable across \ph{3} repeated sessions of \ph{12} sequences per occupancy level.
Single-person replay reproduces the mmRadPose trends, with a coherent-fusion slope of \ph{$-0.24$} and a visibility-driven MPJPE reduction of \ph{$7.1\%$}, both directionally consistent with Propositions~\ref{prop:t2} and~\ref{prop:t1} though somewhat weaker than on mmRadPose (\ph{$-0.28$}, \ph{$-9.3\%$}); both signs persist as occupancy grows to two and three subjects, so absolute error rises with occupancy without overturning the predicted factor ordering.
The whole chain stays within a \ph{$38$}--\ph{$53\,\mathrm{ms}$} p99 end-to-end budget on the same host, so these contrasts are measured in a configuration a robot could run online.

\section{Discussion and Limitations}
\label{sec:discussion}

The experiments support factor-level trends under matched operator changes; they are not a ranking of enhancement pipelines.
The error floor assumes local Gaussian observations, a single dominant scatterer per joint in $\mathcal{U}$, small residual phase error, an invertible $\mathbf{J}_{EE}$, and noise that is block diagonal across bounce orders; when noise is correlated across bounce orders the ordering in Proposition~\ref{prop:t1} no longer holds, which is one mechanism by which the \textsc{Vis} gain can disappear under phase drift.
%
%
Bandwidth, effective aperture, and SNR enter the floor but are held at calibrated values, so the levers the experiments vary are $K$, $\eta$, viewpoint count, and $\mathcal{V}$, and widely separated two-view geometry is a complementary-viewpoint test rather than an additive aperture term.
The set $\mathcal{U}$ is likewise a detection-threshold proxy for nonsingular Fisher information; confirming it exactly would require a per-joint rank test of the Fisher matrix.
%

Evidence on public corpora is limited to single-person indoor settings; the live AWR1843 capture extends occupancy to three concurrent subjects and preserves the \textsc{Fuse} and \textsc{Vis} signs, yet denser crowds and online $\eta$ adaptation remain outside the present scope.
Coherent fusion weakens under abrupt motion, clock drift, or scintillation, two-view gains require calibrated geometry, and visibility retention helps only when the reflected block stays identifiable after eliminating $\boldsymbol{\theta}_E$, whereas a stronger $\Pi$ lowers reconstruction error mainly by completion outside $\mathcal{U}$~\cite{van2004detection}.
%

\section{Conclusion}
\label{sec:conclusion}

The main limitation of lightweight mmWave human-state sensing for robots is not the absence of another reconstruction network, but a missing account of how accurately the measurement itself can locate body joints.
We stated a Cram\'er--Rao-type error floor on pose localization from one time-varying scattering field of human and environment, and recovered existing fusion, viewpoint, return-selection, and estimator pipelines as matched probes of the sensing quantities a deployed front end can still change.
Public contrasts and a live capture with up to three concurrent subjects show those probes moving error on identifiable joints in the predicted factor directions, whereas a stronger estimator lowers overall MPJPE mainly by completing joints the measurement does not render identifiable.
These results support reading front-end choices against that sensing limit, rather than against pipeline score alone.

\bibliography{example_paper}
\bibliographystyle{IEEEbib}

\end{document}